# Optimization Techniques for Sentiment Analysis Based on LLM (GPT-3)

Tong Zhan[1*], Chenxi Shi[1.2], Yadong Shi[1.3], Huixiang Li[2], Yiyu Lin[3]

[1*] Computer Science ,Columbia University,NY, USA
[1.2] Telecommunication Systems Management,Northeastern University,Boston, MA, USA
[1.3] Computer Science, Fudan University, ShangHai, China
[2] Information Studies, Trine University, AZ, USA
[3] Computer Science and Engineering, Santa Clara University, CA, USA

**Abstract.**

With the rapid development of natural language processing (NLP) technology, large-scale pre-trained language models such as GPT-3 have become a popular research object in NLP field. This paper aims to explore sentiment analysis optimization techniques based on large pre-trained language models such as GPT-3 to improve model performance and effect and further promote the development of natural language processing (NLP). By introducing the importance of sentiment analysis and the limitations of traditional methods, GPT-3 and Fine-tuning techniques are introduced in this paper, and their applications in sentiment analysis are explained in detail. The experimental results show that the Fine-tuning technique can optimize GPT-3 model and obtain good performance in sentiment analysis task. This study provides an important reference for future sentiment analysis using large-scale language models.

## 1. Introduction

Natural language processing (NLP), an important branch of the field of artificial intelligence, aims to enable computers to understand and generate natural language used by humans. With the rapid development of technology, [1]NLP has penetrated into every area of our lives, from intelligent voice assistants to online translation tools to sentiment analysis on social media, its application is everywhere. Sentiment analysis, as one of the important applications of NLP, aims to understand human emotions and attitudes by analyzing emotional colors in text. Humans express emotions in a variety of ways, including text, speech, images, etc. Sentiment analysis focuses on emotional judgment and classification of text data. Sentiment analysis is widely used in social media monitoring, public opinion analysis, product reviews and other fields, which can help enterprises understand users' emotional tendencies, so as to adjust marketing strategies or improve product quality.

However, [2]traditional sentiment analysis methods often rely on hand-designed features and rules, which are difficult to capture complex emotional information in text, and have limited adaptability to different contexts and cultural backgrounds. In recent years, with the development of deep learning and other technologies, sentiment analysis methods based on large pre-trained language models (such as GPT-3)[3] have gradually become a research hotspot. By learning language patterns and semantic information in large-scale text data, these models can automatically learn and capture emotional features in text, so as to achieve more accurate and intelligent sentiment analysis. This paper aims to explore GPT-3-based sentiment analysis optimization technology to improve model performance and effect, and further promote the development of NLP field.

## 2. Formatting the title, authors and affiliations

*2.1. Traditional Natural Language Processing (NLP)*

Natural language processing (NLP) is a field of AI that aims to enable computers to understand and use human language in order to perform useful tasks. Natural language processing is divided into two parts: Natural language understanding[4] (NLU) and Natural language generation (NLG). NLP is an area of AI responsible for understanding and processing human language. NLP is part of the overlap between AI, computer science, and linguistics, where the main goal is to make computers understand statements or words expressed in human language.

NLP can be used to solve many problems. Because the amount of data available for text data is very large, it is impossible for people to process all of it. Wikipedia averages 547 new articles per day, and more than 5 million articles in total. Obviously, one person can't read that much information. NLP faces three challenges: collecting data, classifying data, and extracting relevant information.

NLP can handle many tedious tasks, such as spam detection, parts-of-speech tagging, and named entity recognition. Using deep learning, NLP can also solve the speech-to-text problem. Although NLP has shown great capabilities, there are no good solutions for problems such as human-machine dialogue, question-answering systems, automatic summarization, and machine translation.

Natural language processing (NLP) is widely used in various fields in practical applications, of which intelligent assistant is a notable representative. Intelligent assistants such as Apple's Siri and Microsoft's Xiaoice are able to answer users' questions and provide services through voice commands and conversations, greatly improving user convenience. [5]In addition, voice input methods such as Sogou input method and Baidu input method also play a vital role in them, they can convert speech into text, greatly improving the user's input efficiency. In addition, natural language processing is also widely used in cross-border e-commerce, tourism translation, social media and other fields. However, natural language processing faces a number of technical challenges, the most important of which is the diversity of speech signals. The speech characteristics of different people are different, and the environmental noise will also cause interference to the recognition of speech signals. Secondly, there is the problem of accent and dialect, which vary greatly in different regions and different cultural backgrounds, which brings troubles to natural language processing. In addition, long speech recognition is also a challenge, because the processing of long speech requires higher computing resources and more complex algorithms.

*2.2. Sentiment analysis*

Sentiment analysis is an important application in the field of artificial intelligence, which can help us understand the emotional color of a piece of text. For example, we can use sentiment analysis to assess the positive or negative tendency of a news report, or understand the public's attitude and emotion towards a certain topic through comments on social media, or extract opinions, analyze themes and dig emotions from text information such as product reviews and movie reviews.

(a) A rules-based approach

Rule-based approach is one of the earliest methods to be applied to sentiment analysis, which is mainly to identify emotional information in text by manually defining a series of rules. Firstly, rules can be designed flexibly according to specific application scenarios, and a large amount of annotation data is not required. However, due to the variety of emotional expression forms, the design of rules needs to take into account different emotional expression ways, and this method is not as robust and adaptable as other methods.

(b) An approach based on sentiment dictionaries

The approach based on emotion dictionary is to use emotion dictionary to identify emotion information in text. An emotion dictionary is a dictionary of emotion words in which each word is labeled positive, negative, or neutral, so that the emotion tendency of the text can be obtained by counting the number and polarity of emotion words in the text.

This method can perform sentiment analysis quickly and is relatively easy to implement. However, the construction of the emotion dictionary requires a lot of manual annotation, and the emotion dictionary may not cover all the expressions of emotion, and the recognition accuracy is also affected by the quality of the emotion dictionary.

(c) Traditional machine learning methods

Traditional machine learning methods include naive Bayes, support vector machines (SVM) and random forest algorithms. The basic idea of these methods is to use the existing annotation data to train the classifier, and then use the trained classifier to perform sentiment analysis on the new text. In this way, although a large number of labeled data can be used for training, thereby improving the performance of the classifier. However, features need to be designed manually, and for different tasks and data sets, features need to be redesigned, and the training process is time-consuming.

(d) Deep learning methods

Deep learning is currently one of the best performing methods in the field of sentiment analysis, it can automatically learn from the original text representation features, and can handle different lengths of text. Deep learning methods include models such as recurrent neural networks (RNN), short term memory networks (LSTM), and convolutional neural networks (CNN). Among them, RNNS and LSTMS can deal with sequence data well, while CNNS can deal with local information in text effectively. These models typically include an embedding layer to translate the text into a vector representation, one or more hidden layers to extract features, and finally an output layer to predict the emotional polarity of the text.

Deep learning sentiment analysis can automatically learn the features in the text without manual feature engineering, and can also use a large number of unlabeled data for training, thereby improving the generalization ability of the model. However, most of the time, a large amount of labeled data is needed to train the model, and the training process is complicated and time-consuming.

*2.3. Large language model GPT-3*

Large language model (LLMs) is a special class of pre-trained language model (PLMs), which is obtained by expanding the model size, pre-trained corpus and computational power. LLMs exhibit special capabilities due to their enormous size and pre-training on large amounts of text data, allowing them to achieve excellent performance in many natural language processing tasks without any task-specific training. The era of [6]LLMs began with [7]OpenAI's GPT-3 model and grew exponentially in popularity after the introduction of models such as ChatGPT and GPT4. We refer to GPT-3 and its successor OpenAI models (including ChatGPT and GPT4) [8]as the GPT-3 Family of Large Language Models (GLLMs). With the growing popularity of GLLMs, especially in the research community, there is an urgent need for a comprehensive review that summarizes recent research advances across multiple dimensions to provide insights into future research directions. In our review paper, we first introduce basic concepts such as transformers, transfer learning, self-supervised learning, pre-trained language models, and large language models. We then give a brief overview of GLLMs and discuss GLLMs performance in various downstream tasks, specific domains, and multiple languages. We also discuss the data annotation and data enhancement capabilities of GLLMs, the robustness of GLLMs, the effectiveness of GLLMs as an evaluator, and finally, summarize several insightful future research directions. In conclusion, this comprehensive review paper will be a good resource for those in academia and industry to learn about the latest research related to the GPT-3 family of large language models. Indexed terms - Large Language Model, GPT-3, ChatGPT, GPT-4, transformer.

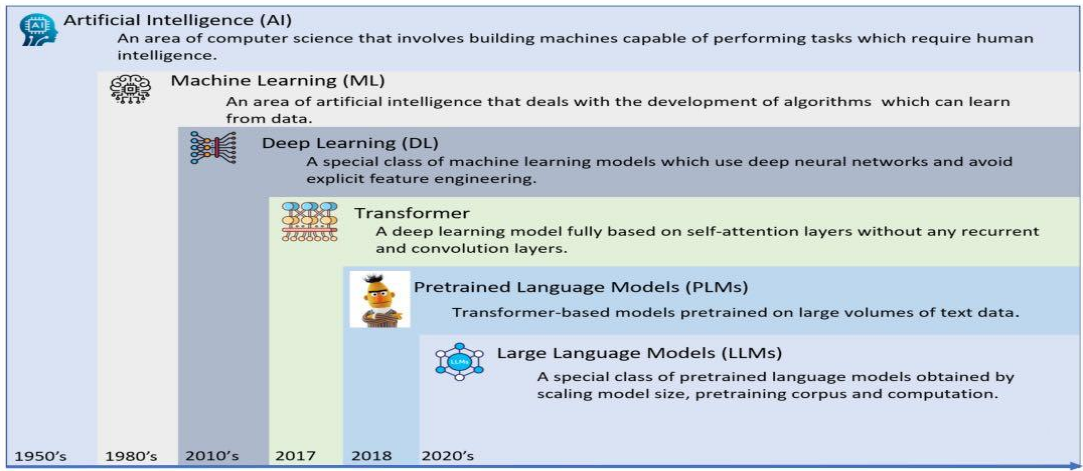

**Figure 1.** Evolution history of large language models

LLMs leverages contextual learning (ICL), a new learning paradigm that does not require task-specific fine-tuning and a large number of labeled instances [9]. LLMs treats any NLP task as a conditional text generation problem and generates the required text output based solely on input prompts, including a task description, test input, and optionally some examples. Figure 1 shows the evolution of AI from machine learning to large language models.

In this section, this project delves into the realm of natural language processing (NLP), emphasizing its role in empowering this text to understand and manipulate human language. From grappling with vast text data to discerning emotional nuances through sentiment analysis, various methodologies are explored, including rule-based approaches and deep learning techniques like recurrent neural networks (RNNs) and convolutional neural networks (CNNs). Moreover, the transformative impact of Large Language Models (LLMs) such as GPT-3 is highlighted, promising to revolutionize sentiment analysis and reshape the landscape of NLP within this text.

## 3. Methodology

### 3.1. GPT-3 model

GPT-3 Language Models are Few Shot Learners Grand model is a very large scale language model developed by OpenAI[10]. The GPT-3 Grand model has a staggering parameter scale of 175 billion parameters, making it one of the largest models of its kind. This superscale gives it powerful language processing capabilities to generate high-quality, coherent text across a wide range of contexts and situations. The GPT-3 big model is based on the Transformer architecture, which is a very popular architecture in natural language processing. It uses a self-attention mechanism, which enables the model to automatically associate information from different locations and take context into account when generating text. This mechanism helps the model better understand and generate text that conforms to syntactic and semantic rules.

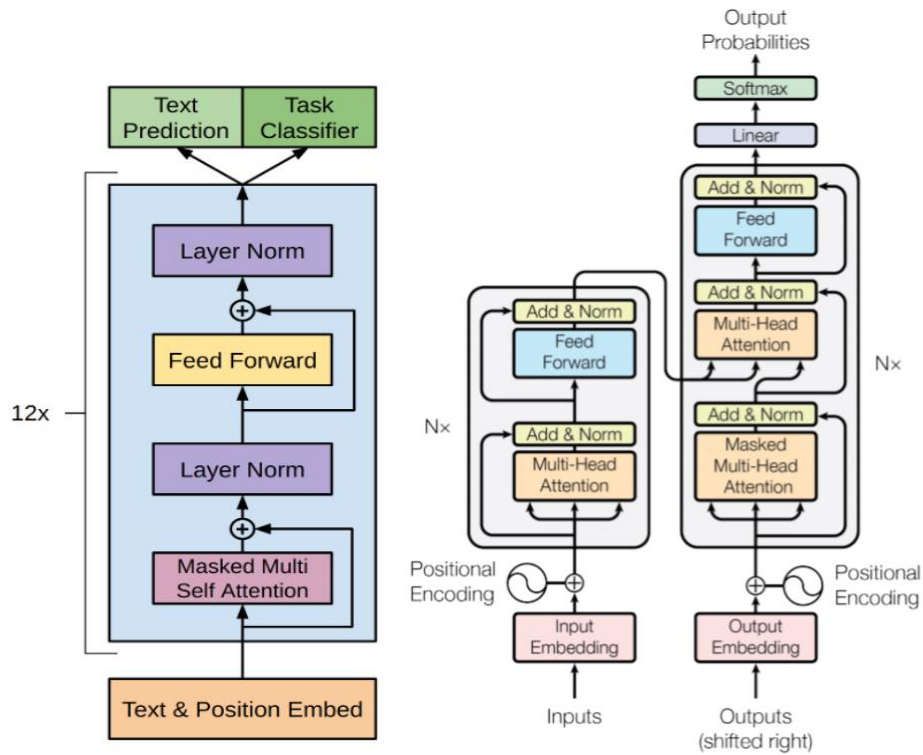

**Figure 2.** The Transformer - model architecture.

In Figure 2, we describe in detail the basis of GPT-3's network architecture. GPT-3 uses a model architecture called Transformer, which has been hugely successful in the field of natural language processing. The Transformer model consists of multiple stacked self-attention layers, each consisting of two sub-layers: multi-head attention and feedforward neural networks. The self-attention mechanism enables the model to maintain long-distance dependencies when processing sequence data, which is particularly important for natural language processing tasks. In addition, GPT-3 introduces a large number of parameters and a deeper network structure, allowing the model to capture more complex linguistic patterns and semantic information.

To learn more about the network architecture of PGT-3, start with the Transformer model and understand how its self-attention mechanism, multi-head attention, and feedforward neural networks work. You can then dive into specific extensions of GPT-3, including how its larger number of parameters and deeper network structure affect the model's performance and effect. By learning the network structure of GPT-3, we can better understand its advantages and application value in natural language processing tasks.

*3.2. Fine-tuning technique*

The process of fine-tuning is not a simple splicing of a [11]pre-trained model with a domain-specific dataset, but a process of fine tuning. First, we need to select the appropriate hyperparameters to adjust the model, such as learning rate, batch size, and training rounds. The selection of these hyperparameters is critical to the success of fine tuning, as they directly affect the performance of the model on a specific task. The learning rate controls the rate at which model parameters are updated, the batch size affects the number of samples taken for each update, and the training rounds determine how many times the model is trained on the entire data set. By adjusting these hyperparameters properly, the model can converge faster and achieve better performance during fine tuning.

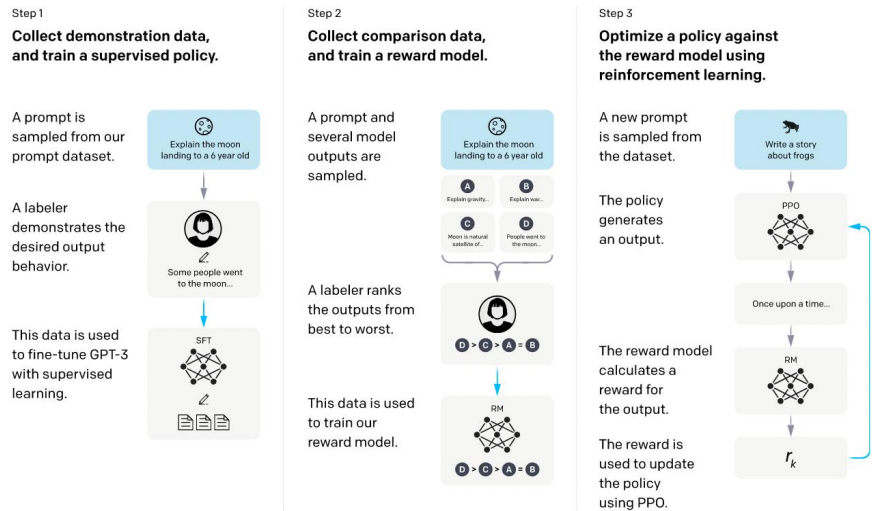

**Figure 3.**ChatGPT-3 large model fine-tuning

In addition, overfitting and underfitting of the model should be considered in the process of fine tuning. Overfitting means that the model performs well on the training set, but poorly on the test set, resulting in insufficient generalization ability of the model. [12-13]Underfitting refers to the inability of the model to fully learn the features and rules of the data set, resulting in poor performance of the model on both the training and test sets. To address these issues, we can mitigate the underfitting problem by preventing overfitting through appropriate regularization techniques such as Dropout and data enhancement methods, while increasing the complexity of the model to improve its expressiveness. By taking these factors into consideration, the fine-tuning process can be optimized effectively, and the performance and generalization ability of the model on specific tasks can be improved.

As can be seen from Figure 3, fine tuning has significant advantages in sentiment analysis technology. First, it can make the pre-trained model better adapt to the needs of specific sentiment analysis tasks, and improve the accuracy and effect of sentiment analysis by adjusting the model parameters accordingly. Second, fine-tuning can help models learn domain-specific language patterns and knowledge, thereby improving the generalization and adaptability of sentiment analysis. In addition, fine-tuning can also solve the domain bias problem of the pre-trained model in the sentiment analysis task, and improve the accuracy and reliability of sentiment analysis by adjusting the model parameters to better adapt to the language characteristics and emotion expression modes of the specific domain. To sum up, fine-tuning plays an important role in sentiment analysis techniques, and the performance and effectiveness of sentiment analysis tasks can be further optimized through appropriate strategies.

*3.3. Application of Transfer Learning*

Transfer Learning (Transfer Learning) is to use the learned and trained model parameters as the starting parameters of the new training model. Transfer learning is a very important and commonly used strategy in deep learning.

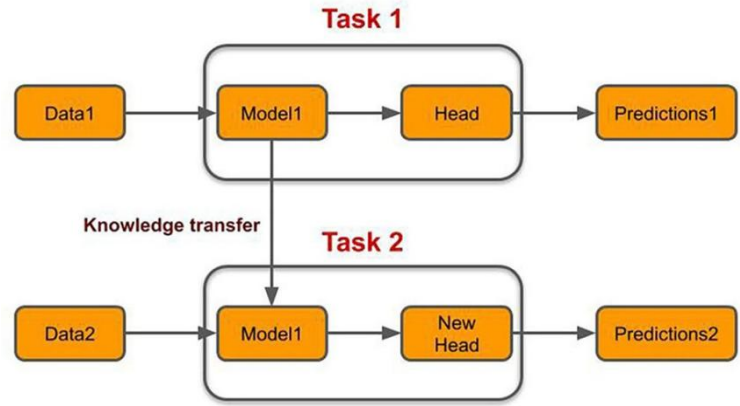

**Figure 5.** Transfer Learning architecture

At present, Transfer Learning is mainly used in the field of computer and causal inference. It focuses on how models learned from one population or task can be transferred to new target populations or scenarios[14].

In computer science, Transfer Learning is A machine learning method that takes the model developed for task A as the initial point and re-uses it in the process of developing the model for Task B. Transfer learning is a new task that improves learning by transferring knowledge from a related task that has already been learned, and while most machine learning algorithms are designed to solve a single task, the development of algorithms that facilitate transfer learning is an ongoing topic of interest in the machine learning community. Sometimes, the observed covariates in the experiment are often less than the number of covariates in the observed target population, so the application of the above method has obvious limitations. In the computer field, there are two very important concepts in transfer learning: [15-16]Domain and Task. domain can be understood as a specific domain at a certain time, for example, book review and TV series review can be regarded as two different domains, and tasks are things to be done, such as sentiment analysis and entity recognition are two different tasks.

*3.4. Sentiment analysis optimization*

GPT-3 has a staggering 175 billion parameters scale and has been pre-trained on a large amount of text on the Internet, giving it powerful language understanding and generation capabilities. Among them, Fine-tuning technology is one of the important functions of GPT-3 and other pre-trained models. Fine-tuning improves the performance and adaptability of the model by training it specifically to the needs of a specific task or domain. While GPT-3 is excellent for small sample learning, the introduction of fine-tuning techniques becomes particularly important when faced with specific needs such as untrained or proprietary data. Through fine tuning, the model can be optimized to meet the needs of specific tasks on the basis of the existing pre-trained model, thus further improving the performance and adaptability of the model. In this paper, we will discuss how to optimize GPT-3 sentiment analysis using fine-tuning techniques to improve the performance and effect of the model on sentiment analysis tasks.

**Common scenarios are:**

Text generation: More accurate and targeted text can be generated by providing relevant data sets and instructional text

Text categorization: Grouping a piece of text into multiple categories, such as email categorization

Sentiment analysis: Analysis of the emotional tendencies of a text, whether positive or negative

This paper will make an attempt at sentiment analysis.

Data preparation and preprocessing are important steps in sentiment analysis model training. First, we obtained online data on hotel reviews, including the content of the reviews and the corresponding emotional labels (1 for positive, 0 for negative). We then need to convert this data into a format acceptable to the GPT model, i.e. {"prompt": "<prompt text>", "completion": "<ideal generated text>"}. To validate and format the data, we use tools provided by openai and operate from the command line. Along the way, we added a suffix separator and a space character at the beginning of the finish to help the model better understand the input data. At the same time, we chose to split the data into training sets and validation sets in order to train and evaluate the model. In the end, the process produces two jsonl files that contain the formatted data and provide reference commands and an estimated time for training.

*3.5. Model result*

**Table 1.** Preprocessing model results

| Model Name | Description | Training / 1K tokens | Inference / 1K tokens |
|---|---|---|---|
| Davinci | The most powerful GPT-3 model capable of completing any task with high quality. | $0.0300 | $0.1200 |
| Curie | Highly capable, faster, and lower cost alternative to Davinci. | $0.0030 | $0.0120 |
| Babbage | Suitable for simple tasks, very fast, and lower cost compared to Curie. | $0.0006 | $0.0024 |
| Ada | Designed for very simple tasks, typically the fastest and lowest cost in the GPT-3 series. | $0.0004 | $0.0016 |

**Analysis of model training results:**

Before training the model, we first add the API key to the environment variable to make calls to the openai API. We then use the command line tools provided by openai to create the fine-tuning task. When creating the fine-tuning task, we specified the file paths for the training set and validation set, and used the --compute_classification_metrics parameter to calculate the model's performance metrics for the classification task, including accuracy, precision, recall, and F1 score. At the same time, we specify the positive class or positive example in the classification task through the --classification_positive_class parameter. However, when executing the command, the model parameters may have been inadvertently not specified, resulting in the Curie model being selected by default for fine tuning. To cancel this training, we used the commands provided by openai to cancel the fine-tuning task. By querying the list of fine-tuning tasks, we found that the model chosen by default is Curie.

Next, we resumed the fine-tuning task and waited for the training to complete. If the training process is interrupted unexpectedly, we can use the commands provided by openai to track the progress of the fine-tuning task. After the training is complete, we can view the performance metrics of the model by querying the results of the fine-tuning task. According to the results, the accuracy of the model is 0.85, which indicates that the model has achieved good performance in the classification task.

*3.6. Experimental discussion*

In the experimental discussion, we should first consider the performance and performance of GPT-3 model and the effect of Fine-tuning for sentiment analysis optimization. The GPT-3 model, with its incredible parameter scale and powerful language processing capabilities, is based on the Transformer architecture and utilizes technologies such as self-attention mechanisms to understand and generate text. Through Fine-tuning technology, we can adjust the model parameters pertinently, make it better adapt to the needs of the emotion analysis task, and improve the performance and effect of the model in the emotion analysis task.

In the experiment, we first pre-processed the hotel review data and converted it into a format acceptable to the GPT model. We then use the tools and command line operations provided by openai to validate and format the data, and split the data into training and validation sets. Then, we optimize

GPT-3 model through Fine-tuning technology for emotion analysis, and conduct training and evaluation. By querying the results of the fine-tuning task, we know that the model has achieved good performance on the sentiment analysis task, with an accuracy of 0.85.

We can then apply the trained model to a real sentiment analysis task. By calling the model API, we can analyze the sentiment of the text and get the prediction result of the model on the sentiment of the text. By analyzing the performance indicators of the model and the practical application effect, we can evaluate the effect of Fine-tuning technology in the optimization of sentiment analysis, and further discuss how to further optimize the performance and effect of the model. Through these experiments and discussions, we can more deeply understand the application potential of GPT-3 model in sentiment analysis tasks, and provide reference and inspiration for future research and application.

**4. Conclusion**

Although GPT-3 has made remarkable achievements in the field of natural language processing, its future development still faces challenges and opportunities. [17]First, as computing resources continue to improve and technology advances, we can expect GPT-3 and its subsequent versions to achieve greater breakthroughs in performance and effects. For example, with the upgrading and optimization of hardware equipment, we can foresee that the efficiency of GPT-3 model in the training and reasoning process will be further improved, so that it can be widely used in more application scenarios.

Secondly, for the optimization and application of GPT-3 model in sentiment analysis tasks, future research and practice will continue to explore more effective optimization techniques and methods. In addition to Fine-tuning, other transfer learning methods, model distillation techniques, etc. can be considered to further improve the performance and effect of the model in the sentiment analysis task. At the same time, the model can be combined with domain knowledge and contextual information to further improve the understanding and analysis ability of text emotion.

In addition, with the continuous development of artificial intelligence technology and the continuous expansion of application scenarios, we can look forward to the application of GPT-3 in more fields. In addition to the field of natural language processing, GPT-3 can also be applied to other fields, such as intelligent decision-making assistance, intelligent customer service systems, intelligent creative assistants, etc., to bring more convenience and innovation to people's life and work.

In general, although there are still some technical challenges and limitations of GPT-3 model, with the continuous progress of technology and the continuous expansion of application scenarios, we have reason to believe that GPT-3 and its subsequent versions will play an increasingly important role in the future, promote the continuous progress of natural language processing technology, and make greater contributions to the development and progress of human society.